\pdfoutput=1

\documentclass[11pt]{article}

\usepackage[]{acl}

\usepackage{times}
\usepackage{latexsym}

\usepackage{enumitem}

\usepackage[T1]{fontenc}

\usepackage[utf8]{inputenc}

\usepackage{booktabs}
\usepackage{subcaption}

\usepackage{microtype}

\usepackage{inconsolata}
\usepackage{graphicx}

\definecolor{blue}{HTML}{4285f4}
\definecolor{lightblue}{HTML}{c9daf8}
\definecolor{darkblue}{HTML}{1c4587}
\definecolor{green}{HTML}{34a853}
\definecolor{lightgreen}{HTML}{d9ead3}
\definecolor{orange}{HTML}{ff9900}
\definecolor{lightorange}{HTML}{fce5cd}
\definecolor{lightred}{HTML}{e06666}
\definecolor{purple}{HTML}{9900ff}
\definecolor{lightpurple}{HTML}{b4a7d6}
\definecolor{gray}{HTML}{cccccc}

\newcommand{\negative}[1]{\textcolor{orange}{ #1}}
\newcommand{\positive}[1]{\textcolor{darkblue}{ #1}}

\usepackage{soul}
\usepackage{todonotes}

\title{Promoting Constructive Deliberation: Reframing for Receptiveness}

\author{Gauri Kambhatla \and Matthew Lease$^*$ \and Ashwin Rajadesingan\thanks{Denotes equal author contributions}\\
        The University of Texas at Austin \\
        \texttt{\{gkambhat, ml, arajades\}@utexas.edu}}

\begin{document}
\maketitle

\begin{abstract}
To promote constructive discussion of controversial topics online, we propose automatic reframing of disagreeing responses to \textit{signal receptiveness} to a preceding comment. Drawing on research from psychology, communications, and linguistics, we identify six strategies for reframing. We automatically reframe replies to comments according to each strategy, using a Reddit dataset. Through human-centered experiments, we find that the replies generated with our framework are perceived to be significantly more receptive than the original replies and a generic receptiveness baseline. We illustrate how transforming receptiveness, a particular social science construct, into a computational framework, can make LLM generations more aligned with human perceptions. We analyze and discuss the implications of our results, and highlight how a tool based on our framework might be used for more teachable and creative content moderation.
\end{abstract}

\section{Introduction}

Constructive deliberation and debate amongst disagreeing views promotes sharing ideas and collective intelligence for innovation, decision-making, and governance \cite{porter-and-schumann-2018}. Such exchanges also promote empathy and tolerance, rather than isolation and polarization \cite{mutz2002cross}. However, online discussion today is often hostile and confrontational, where exposure to opposing views in this setting can instead provoke even greater polarization \cite{bail_2018}. 

\begin{figure}
    \centering
    \includegraphics[width=\linewidth]{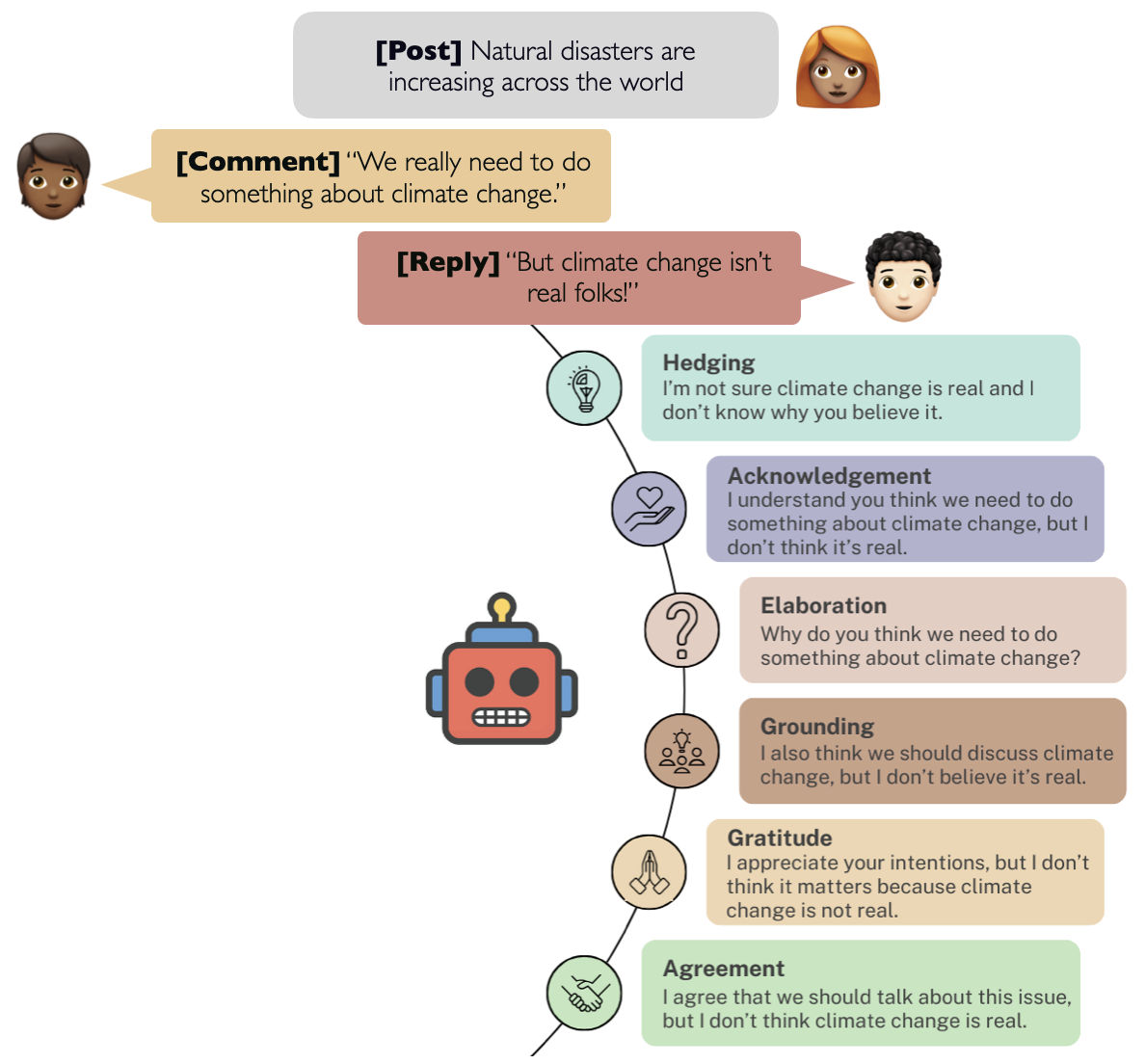}
    \caption{Example of generated receptiveness reframes. Given a post, comment, and reply from Reddit \cite{Pougue-Biyong_undated-as}, we generate a reframe for each of our six receptiveness-signaling strategies.}
    \label{fig:strategy_generation}
    \vspace{-1em}
\end{figure}

Current methods for promoting pro-social discussion and debate online are limited. For example, community guidelines may principally focus on preventing harms rather than promoting pro-social norms. In addition, moderator controls are often fairly coarse: simply blocking users or posts is scalable but rigid, while crafting open-ended messages to guide behavior is extremely flexible but can be difficult to scale \cite{jhaver2019does}. 

NLP research for assisting scalable content moderation has also largely focused on anti-social tasks such as removing toxic speech \cite{pamungkas-etal-2020-really, ranasinghe_2021}, reducing offensive language with style-transfer \cite{atwell-etal-2022-appdia}, and removing content that violates community rules \cite{ye-etal-2023-multilingual}. While some studies have begun to pursue content moderation through the lens of promoting \textit{pro-social} content \cite{ashida-komachi-2022-towards, yu-etal-2022-hate, madaan-etal-2020-politeness, kim-etal-2022-prosocialdialog}, promoting constructive deliberation has received relatively little attention. This is a difficult task, involving nuanced and subtle linguistic cues that underlie productive discussions. 

Research from communications, psychology, and linguistics suggests how we might promote open-mindedness to opposing views. In particular, \citet{yeomans_2020-kj}'s conceptualization of \textit{conversational receptiveness} focuses on conveying openness while still disagreeing. Others have studied how to measure receptiveness towards disagreeing perspectives \cite{Minson2020-vl, Minson2022-gx}. However, there has been limited success in translating such theory into practice. 

In this work, we seek to bridge NLP research on textual \textit{reframing} \cite{Sharma2021-ac,ziems-etal-2022-inducing,madaan-etal-2020-politeness,chakrabarty-etal-2021-entrust} with social science theories on receptiveness. Our aim is to promote pro-social deliberation online; particularly to foster productive discussion for contentious issues. We develop a computational framework to signal receptiveness while preserving the original meaning of reframed content, studying the capabilities of large language models (LLMs) for this task. We illustrate the value of translating social science theory to a computational framework by showing how model generations created with our framework better align with human notions of receptiveness, allowing for more useful content moderation tools.

Models that are more receptive-aligned might allow for content moderation tools that are valuable to users and moderators despite being automated. For moderators, a tool built on our receptiveness framework could allow for more scalable and creative moderation. For users, such a tool might provide learning opportunities, suggesting possible revisions if their post is flagged as unreceptive. Users could learn how to be more receptive from suggestions, whether or not they accept them. 

Here, we operationalize receptiveness through a framework of six lower-level strategies from social science theory to make messages more receptive to opposing viewpoints. We ground this conceptual work in a concrete study of Reddit posts, illustrated in Figure \ref{fig:strategy_generation}. We compile a new corpus of automatically reframed Reddit replies and assess (via automatic measures and human evaluation) the validity of the generated replies.

We compare reframes generated by our framework to original replies and baselines. We find they are significantly more aligned with what humans find receptive, eliciting relatively low negative emotional reactions, high curiosity to engage with opposing views, and reduced belief that the user is trying to shut down discussion. 
We also explore the interaction between toxicity and receptiveness reframing, finding that more toxic content benefits more from reframing. We end with broader discussion of receptiveness for content moderation and mitigating polarization. We create and share a corpus of receptive-reframed replies, baseline reframes, and human annotations of receptiveness (16.6k reframes and over 9.2k total annotations)\footnote{Data is available at: \url{https://github.com/GauriKambhatla/constructive_deliberation}}.

\section{Related Work}
\label{sec:background}
\vspace{-0.5em}

\subsection{Constructive deliberation and receptiveness in communications}
\label{subsec:receptive}

Constructive discussions across differences require that individuals not only voice their own arguments, but that they also meaningfully engage with arguments presented by others \cite{bachtiger2019mapping}. Researchers typically draw on the related concepts of reciprocity, listening and receptiveness to study such reciprocal engagement. 

\citet{graham2003search} view reciprocity as the first step in this process and define it as ``the giving and taking of validity claims, arguments, and critiques among participants.” They argue that reciprocity followed by reflexivity and empathy is required for deliberations to result in mutual understanding. \citet{esau2022creates} further distinguish reciprocity from replying, defining ``a reciprocal comment'' as one ``which is on topic, respectful in tone and reasoned.’’ \citet{scudder2020beyond} instead argues for democratic listening, a communicative act beyond simply listening that is performed ``for the sake of considering what others have to say’’. 

We draw on \citet{yeomans_2020-kj}'s conceptulization of conversational receptiveness as it more directly focuses on receptiveness when expressing disagreement: the ``extent to which participants in disagreement communicate their willingness to engage with each other’s views.’’ Their study identified numerous linguistic markers of receptiveness during disagreement, such as hedging and acknowledgement.

\citet{Minson2020-vl} developed a scale for \textit{dispositional} receptiveness (receptiveness with regard to beliefs or personality, rather than through communication). Items in the scale correspond to four factors: negative emotions, curiosity, bias, and belief of lack of openness. We use these factors to develop our receptiveness index (Section \ref{sec:recindex}).

\subsection{Constructive deliberation and NLP}

There has been some NLP research on constructive deliberation. Recent research focuses on \textit{grounding}, or constructing a shared basis of understanding. \citet{cho-may-2020-grounding} create a dataset of two-turn dialogues with a particular grounding speech act (``yes, and'') that implies grounding. They find that a model fine-tuned with this dataset encourages more grounded conversations. \citet{shaikh-etal-2024-grounding} study the discrepancies between human grounding and LLM generations with grounding, which they call the ``grounding gap''. They come up with a set of \textit{grounding acts} to quantify LLM attempted grounding, similar to our strategies for receptiveness.  

Other work looks at constructive discourse in real-world settings, such as in the space of content moderation. \citet{cho-etal-2024-language} define metrics for conversational moderation effectiveness, and evaluate language models as moderators. They find that language models that are prompted using insights from social science can provide good feedback to users, but struggle to increase users' levels of cooperation and respect. \citet{park-etal-2021-detecting-community} study using community norms for better context-specific automated content moderation. They find that explicitly encoding community norms can allow models to detect community norm violations with high performance.

\subsection{Reframing in NLP}

Other studies ground reframing (style transfer while preserving meaning) techniques in social science theory, with applications to various domains. \citet{ziems-etal-2022-inducing} study \textit{positive reframing}: reframing pessimistic tweets to be more optimistic using strategies from psychology literature. They introduce a conditional generation task for this. \citet{Sharma2023-jd} likewise look at reframing negative thoughts, but draw from \textit{cognitive reframing} techniques. They use a retrieval-based in-context learning approach to reframing. \citet{chakrabarty-etal-2021-entrust} study reframing arguments to be more trustworthy (defined as non-partisan and without appeals to fear) with arguments on the subreddit ChangeMyView. \citet{madaan-etal-2020-politeness} introduce politeness transfer, or converting to polite text using a two-stage tag and transfer approach to reframing. 

Closest to our work, \citet{Argyle2023-mz} seek to improve political discussions using GPT-3 to reframe conversations
with opposing views in terms of restatement, validation, and politeness. They find that participants in chats with GPT-reframing assistance reported higher conversational quality and democratic reciprocity than those without assistance. Our work also pursues reframing, but we focus on grounding and translating a specific theoretical construct -- conversational receptiveness -- into practice. Furthermore, we extend our analyses to better understand individual factors that influence perception of receptiveness, and evaluate contexts in which reframing is most effective.

\section{Approach}
\label{sec:approach}

Our broad goal is to foster pro-social online discussions in which participants constructively engage with others having opposing viewpoints. To this end, we pursue automated techniques to {\em reframe} online discussions, i.e., rephrasing messages in a manner that would preserve a message's original meaning while modifying its tone to convey greater openness to opposing viewpoints. 

In Section \ref{sec:strategies}, we introduce and motivate six specific methods we apply for conveying receptiveness curated from social science literature that we transform into reframing strategies. These are also illustrated in Figure \ref{fig:strategy_generation}. Following this, we describe our specific task formulation and data used (Section \ref{sec:task}). Finally, we describe our method for automating this reframing (Section \ref{sec:machine}).  

\subsection{Reframing strategies for receptiveness}
\label{sec:strategies}

Brown and Levinson (\citeyear{brown1987politeness})'s politeness theory suggests that people desire in social interactions to be appreciated and maintain a positive self image (``positive face'') and to not be curtailed or imposed upon by others (``negative face''). Disagreements may cause disrespect or offense, threatening positive face, and/or may intrude on the receiver's autonomy, threatening negative face. Face threatening disagreements are especially evident when discussing politics on social media where partisan distrust \cite{iyengar2019origins} coupled with the limited affordances available to gauge perceptions \cite{bail2022breaking} may derail even well-intended interactions. Numerous strategies have been identified to mitigate perceptions of threats to face and continue the conversation while still expressing disagreement \cite{Minson2022-gx,yeomans_2020-kj}. 

In this study, we focus on six key strategies (described below) to signal receptiveness, which we pursue in our automated reframing work.

\paragraph{Hedges} are words or phrases that add ambiguity in a way that softens the force of a statement, thus potentially saving negative face of the receiver. They signal possibility in arguments, allowing writers to present ``their claims with caution, and enter into dialogue with their audiences'' \cite{hyland1996talking}. Although hedges may weaken the claim, they tend to strengthen the argument as hedged claims are harder to falsify \cite{georg1997hedging}. Prior work identifies hedges as a potential signal for receptiveness \cite{Minson2022-gx,yeomans_2020-kj}.  Examples of hedges: "may", "perhaps" and "I think".

\paragraph{Elaboration} includes asking for further details, clarifying understanding, and repeating what was heard. Seeking more details may be viewed as delaying expressing disagreement, potentially softening threats to negative face \cite{Pomerantz_2021}. Further asking for elaboration signals expressed interest in the speaker's view which is known to increase perceptions of receptiveness \cite{Chen2010-je}. Examples of elaboration: "what I heard you say was that...", "what did you mean by...".

\paragraph{Grounding} is often an essential precondition for conflict resolution \cite{van2020common}. Highlighting and making explicit shared beliefs, assumptions and knowledge is known to increase conversational receptiveness in civic-minded workshops to reduce polarization \cite{Oliver-Blackburn2023-bm, listen-first-project}. \textit{Acknowledgement} and \textit{agreement} (strategies discussed below) can be viewed as different grounding actions \cite{Clark1989-cu, Traum1999-bn}. We chose to separate these into individual strategies as they may result in different levels or degrees of groundedness which may impact receptiveness. Our expectation is that highlighting specific consensus (agreement strategy) or using affirming speech acts (acknowledgement strategy) may signal receptiveness in a different manner than emphasizing broader common ground. Examples of grounding: "we both believe...", "we both think...".  

\paragraph{Acknowledgement} conveys understanding and validate another’s feelings, appealing to their positive face. Such affirming speech acts signify to the receiver that the speaker is engaging in good faith, likely resulting in reciprocation \cite{Rajadesingan2021-pv}. While \textit{acknowledgement} simply conveys understanding of the speaker's point of view, \textit{grounding} emphasizes common ground and commonalities between the discussants' viewpoints.  Examples of acknowledgement: "I understand what you mean", "I see where you're coming from".

\paragraph{Agreement} typically takes the `yes,.. but' form \cite{Pomerantz_2021}. This mitigation strategy allows individuals to delay and minimize the force of disagreement with an explicit but nominal token of agreement. While \textit{agreement} simply makes explicit some agreement about the speaker's statements, \textit{grounding} also includes highlighting even broader shared beliefs. Examples of agreement: ``I am also a libertarian but I disagree on this \ldots'').

\paragraph{Expressing Gratitude} often yields more positive perceptions and increased trustworthiness of the speaker \cite{percival2020say,dunn2005feeling}. Examples of gratitude: "thank you for your comment", "I appreciate you bringing this up".

\subsection{Task formulation and data}
\label{sec:task}

\paragraph{Task.}  To ground our study, we investigate automated reframing of Reddit discussions. These discussions begin with a post in which someone typically shares a link to news or other content. Next, users comment on the post, after which other users reply to these comments. This initial comment-reply plays an important role in establishing the tone and basis for any ensuing discussion, so we focus our reframing efforts here. In particular, could the reply be reframed to convey greater openness to the opposing viewpoint in the comment while preserving its original meaning?

\paragraph{Data.} English Reddit data is drawn from the Debagreement dataset \cite{Pougue-Biyong_undated-as} for disagreement detection. It consists of comments and replies from controversial subreddits: r/Brexit, r/democrats, r/Republicans, r/climate, and r/BlackLivesMatters. We exclude r/Brexit due to our US-based annotators being less familiar with Brexit-related people and issues. We select examples that are labeled as disagreement and where the comment and reply are at most 30 words each, since preliminary pilots revealed that longer text increased the cognitive burden on annotators. The average word length of the comments and replies in the Reddit data are 36.15 words and 33.12 words respectively, indicating our length-constrained data is not much smaller than average. 

We run all replies through the Perspective API\footnote{\url{https://perspectiveapi.com/}} and exclude any reply with predicted toxicity above 0.9, as these were typically too toxic to be reframed without significantly altering their original meaning. Examples of replies for varying toxicity levels is shown in Appendix Table \ref{tab:toxicity_examples}. Further discussion of the interaction between toxicity and reframing is explored in Section \ref{sec:receptiveness_experiment}).

\subsection{Generating reframes}
\label{sec:machine}

To generate reframes for each reply, we use \texttt{gpt-4} \cite{openai2023gpt4}. To improve generation quality, we use in-context learning. Specifically, we provide five example comment-reply-reframe triples with each instruction prompt. The five comment-reply pairs are drawn from the Reddit data and held constant across the six reframing strategies. For each strategy and for each of the comment-reply pairs, we write a reframe of the reply for the strategy. This requires a total of 30 reframes over the six strategies and five comment-reply pairs. 

\paragraph{Choosing in-context examples.} We choose one comment-reply pair from each subreddit, and reframed responses were written by one of the authors. The final reframes are the result of multiple iterations with feedback from the other authors. In this work, we selected and collectively reviewed real comments from each subreddit for use as our in-context examples. Future work might employ a more calculated strategy to choose examples that better delineate the different strategies.

\paragraph{Baseline reframes.} We consider two zero-shot prompts with \texttt{gpt-4} as baselines. For the \textit{paraphrase baseline}, we prompt the model to simply paraphrase the reply for each comment-reply pair. This baseline serves to distinguish whether observed differences stem from our specific reframing strategies or any language model generation. For the \textit{receptiveness baseline}, we directly prompt for receptiveness, without reference to any of our six strategies. This baseline allows us to evaluate our social science-informed reframing framework against a generic prompt to reframe for receptiveness.

In total, we generate six reframes (plus two baseline reframes) for each of $\sim$2k comment-reply pairs in the subset of the Debagreement dataset that is within our constraints (within the four subreddits, under the 30-word maximum, with a toxicity score under 0.9, and where the comment and reply disagree), yielding 16.6k generations. Appendix \ref{appendix:prompts_and_examples} lists full prompts and few-shot examples.

\section{Validation Experiments}
\label{sec:validation}
\vspace{-0.5em}
We next present analyses conducted to validate our reframe generation methods. We assess whether our reframes satisfy three key properties: 1) demonstrate the six distinct strategies for receptiveness; 2) preserve the meaning of original replies being reframed; and 3) are contextually relevant to the original comment.

\begin{table}[]
    \small
    \centering
    \begin{tabular}{ p{0.2\linewidth} p{0.31\linewidth} p{0.1\linewidth} p{0.1\linewidth}}
    \toprule
    {\bf Strategy} & {\bf $\Delta$ Trigrams} & {\bf $P(t|s)$} & {\bf $P(s|t)$} \\
    \midrule
    Hedging & \textit{\small it might be} & 0.062 & 0.703\\
    & \textit{\small might not be} &  0.037 & 0.393\\
    & \textit{\small it seems like} & 0.027 & 0.271\\
    \midrule
    Acknowledg. & \textit{\small I understand your} & 0.061 & 0.397\\
    & \textit{\small your point about} & 0.023 & 0.371\\
    & \textit{\small I see your} & 0.022 & 0.382\\
    \midrule
    Elaboration & \textit{\small it sounds like} & 0.086 & 0.807\\
    & \textit{\small sounds like you're} &  0.036 & 0.940\\
    & \textit{\small are you suggesting} & 0.022 & 0.816\\
    \midrule
    Grounding & \textit{\small I also think} & 0.058 & 0.955\\
    & \textit{\small I agree that} & 0.025 & 0.237\\
    & \textit{\small I also believe} & 0.020 & 0.977\\
    \midrule
    Gratitude & \textit{\small thank you for} & 0.170 & 0.986\\
    & \textit{\small you for your} & 0.068 & 0.985\\
    & \textit{\small for sharing your} &  0.043 & 0.992\\
    \midrule
    Agreement & \textit{\small I agree that} & 0.080 & 0.748\\
    & \textit{\small I understand your} &  0.067 & 0.381\\
    & \textit{\small I see your} & 0.029 & 0.450\\
    \bottomrule
    \end{tabular}
    \caption{Most common trigrams \textit{added} to reframes vs.\ original replies. Trigrams are ordered by $P(t|s)$. $P(t|s)$ denotes probability of the trigram given the strategy, and $P(s|t)$ the probability of the strategy given the trigram.}
    \label{tab:qualitative_trigrams_added}
    \vspace{-1em}
\end{table}

\subsection{Do reframes reflect the six strategies?} 
\label{sec:validation_strategies}
We inspect differences in the most common trigrams found in generated reframes vs.\ the original replies.
We study two probability distributions: (a) $P(t|s)$, the probability of a trigram $t$ being generated by a given strategy $s$, and (b) $P(s|t)$, the probability of given observed trigram $t$,  having been generated by a strategy $s$. The former tells us whether the model is adding appropriate linguistic phrases for each strategy, while the latter measures how distinct the generations are for each strategy.  

The top trigrams and probabilities are listed in Table~\ref{tab:qualitative_trigrams_added}. For each strategy, trigrams are shown ordered by $P(t|s)$. Regarding (a), we see that the most common trigrams being added do appear to be consistent with each  strategy. Further, except for the ``thank you for'' trigram, trigrams appear less than 10\% of the time as part of each strategy, implying that there is reasonable variety in the way the strategy is expressed. In regard to (b), we find that the majority of the strategies are unique. However, the top trigrams in {\em agreement} overlap with {\em grounding} and {\em acknowledgement}. Recall that agreement represents a narrower subset of grounding; while agreement simply makes explicit any agreement with the speaker’s statements, {\em grounding} also includes highlighting broad shared beliefs. Further, {\em acknowledgement} conveys understanding of the speaker's statement without necessarily agreeing with them, and is often considerd an aspect of grounding. As a result, the overlap between the three strategies is not surprising, though suggests that our expectations for differentiation (discussed in Section \ref{sec:strategies}) did not bear out. This merits further investigation into alternative formulations of these strategies as well as an exploration of others. We discuss more about this issue in \textit{Limitations}. The amount of trigram overlap between the different strategies is shown in Appendix Figure \ref{fig:overlap}.

\subsection{Do the reframes preserve meaning?}
\label{sec:validation_meaning}

We assess the degree to which generated reframes preserve the meaning of original replies with three measures. First, we use {\em form similarity} to measure the average number of distinct n-grams for n=\{1, 2, 3, 4\} per prior work \citep{Tevet2021-nm}. Second, we consider {\em semantic similarity} via SBERT scores \citep{reimers-gurevych-2019-sentence}. Third, we measure {the ratio of contradictions} to non-contradictions between the original and reframed reply using an out-of-the-box DeBERTa model trained on MNLI and SNLI\footnote{\scriptsize \url{https://huggingface.co/cross-encoder/nli-deberta-v3-large}}. 
Following \citet{yerukola2023contextRewrite}, we perform automatic evaluations for form and semantic similarities between the reframe and concatenation of the original reply and the original comment, which provides context for the reply. For contradictions, we only seek to determine whether the generated reply contradicts the original, so we exclude the original comment.

{Table \ref{tab:automatic_validation_meaning}} shows that our reframes
exceed both baselines in both form and semantic similarity with the original reply concatenated with context. When we compare our reframes to the original reply without the context, both baselines' form and semantic similarity scores are higher than our reframes. This indicates that our strategies incorporate context, which has been shown to be greatly preferred \citep{yerukola2023contextRewrite}. 
While the paraphrase baseline exhibits the lowest ratio of contradictions, all strategies are very low (< 3\%); generated replies rarely contradict the original, as desired. We note that these methods do not consider potential added hallucinated content; we are only checking if existing content is retained.

\begin{table}[!t]
\small
    \centering
    \begin{tabular}{c | c c c}
    \toprule
    \textbf{Strategy} & \textbf{Distinct} & \textbf{SBERT}  & \textbf{PC ($\downarrow$)}\\ & \textbf{Ngrams} ($\downarrow$) & \textbf{Score} ($\uparrow$) &  \\
    \midrule
    Paraphrase (B1) & 0.54 & 0.59 & 0.010\\
    Receptive (B2) & 0.56 & 0.56 & 0.050\\ 
    \midrule
    Hedging & 0.46 & 0.61 & 0.013\\
    Acknowledge & 0.47 & 0.71 & 0.024\\
    Elaboration & 0.44 & 0.76 & 0.016\\
    Grounding & 0.45 & 0.70 & 0.019\\
    Gratitude & 0.47 & 0.64 & 0.028\\
    Agreement & 0.43 & 0.75 & 0.025\\
    \bottomrule
    \end{tabular}
    \caption{Automatic validation of meaning preservation of model generations. Distinct ngrams and SBERT score are taken between original dialogue (comment + reply) and reframed reply. PC indicates the proportion of contradictions between original and reframed replies. All values lie in [0,1]. $\downarrow$ / $\uparrow$ arrows mean  smaller/larger values are better. ``B'' indicates a baseline.}
    \label{tab:automatic_validation_meaning}
\end{table}

\subsection{Are the reframes contextually relevant?}
\label{sec:validation_plausibility}
We evaluate whether our reframing framework produces replies that are contextually relevant to the comment. For this analysis, we compare a subset of our reframes with the original reply from Reddit. We ask annotators whether each reply shown constitutes a reasonable response \textit{to the original comment} using a 5-point Likert scale (0 = reply doesn't make sense, 4 = the reply is very reasonable) to determine whether annotators think a reply is relevant in context of the comment. The interface is shown in Appendix Figure \ref{fig:plausibility-interface}. 
Annotations are collected via Amazon Mechanical Turk (MTurk). We describe the steps taken to ensure quality annotation in Appendix \ref{appendix:quality_annotation}. We give annotators 420 examples in multiple batches, and we assign 3-7 workers per example (3.7 on average). Reasonability scores are averaged across annotators and examples within a strategy. {Table \ref{tab:plausibility_scores}} shows results.

\begin{table}[!t]
    \centering
    \begin{tabular}{ll}
        \toprule
         \textbf{Strategy}&  \textbf{Reasonability}\\
         \midrule
         Original& 2.32 $\pm$0.150\\
         \hline
         Hedging& 2.77 $\pm$0.238\\
         Acknowledgement& 3.33 $\pm$0.256\\
         Elaboration& 3.04 $\pm$0.255\\
         Grounding& 3.24 $\pm$0.500\\
         Gratitude& 3.05 $\pm$0.285\\
         Agreement& 3.17 $\pm$0.268\\
         \bottomrule
    \end{tabular}
    \caption{Mean reasonability scores for each strategy, with 95\% CI. Appendix Table \ref{tab:t_test_statistic_plausibility} shows paired t-test for differences between each strategy and the original. }
    \label{tab:plausibility_scores}
    \vspace{-1em}
\end{table}

We conduct a paired t-test between the examples for each strategy and the original to determine if the difference is statistically significant. We find that the mean reasonability score for each of the strategies is significantly higher than the reasonability of the original reply (see Appendix Table \ref{tab:t_test_statistic_plausibility} for mean differences, test statistics, and p-values). This indicates that our reframes are perceived to be reasonable replies to the original comments. 

Surprisingly, the original, human-written replies are perceived to be significantly less reasonable, likely because some of the original replies do not directly respond to the comment (which is common in social media), while the generated replies almost always take the context of the original comment into account. See Appendix Table \ref{tab:original-replies-bad} for examples. Another possible reason is that annotators may have interpreted \textit{reasonable response} to mean something more akin to \textit{receptive response}. We use ``reasonability'' as a derivative of ``plausibility'', which was used in prior work \cite{zhou-etal-2023-cobra} for a similar generation validation experiment. We found during pilot studies that ``reasonability'' was more intuitive than ``plausibility'' to annotators.

\section{Receptiveness Experiment}
\label{sec:receptiveness_experiment}
\vspace{-0.5em}
\subsection{Annotation task}
We conduct an experiment on MTurk to evaluate whether the generated reframes are perceived to be more receptive than the original replies. Appendix \ref{appendix:quality_annotation} describes how we filter the MTurk workforce to ensure quality of collected data. 

As discussed in Section \ref{sec:task}, our data consists of comment-reply pairs, along with generated reframes for two baselines and our receptiveness strategies. We use 9 versions of each of the 75 original replies: the original, the two baselines, and the six receptive reframes. Annotators are asked to compare the relative receptiveness between two alternative replies to a comment: the original vs.\ one of our reframes or a baseline generation. Annotation interfaces are shown in Figures \ref{fig:receptiveness-interface-1}, \ref{fig:receptiveness-interface-2}, and \ref{fig:receptiveness-interface-3} in Appendix \ref{appendix:interfaces}. Annotation is performed in multiple batches with 3-6 workers per example (4.4 on average). We use Krippendorff's $\alpha$ for interannotator agreement, and get a score of 0.54, which makes sense given the subjective nature of this annotation task. 
We study the 95\% confidence interval around the receptiveness scores to evaluate how this disagreement affects variance around the mean receptiveness scores. To evaluate the reframes, we develop a receptiveness index translated from social science literature, as described below.

\subsection{Dependent variable: Receptiveness index} \label{sec:recindex}
We translate \citet{Minson2020-vl}'s {receptiveness index}  for \textit{dispositional} receptiveness to \textit{conversational} receptiveness. Their index was quantified via an 18 dispositional statement scale. Based on responses, they surfaced a set of four factors: 
\begin{enumerate}[topsep=0.5em, itemsep=-0.25em, leftmargin=*]
    \item \textbf{F1 Emotion} [\textit{reverse-coded}]: negative emotional reactions towards opposing (attitude-incongruent) views
    \item \textbf{F2 Curiosity}: intellectual curiosity one might have towards opposing views (desire for greater insight/information about the beliefs of others) 
    \item \textbf{F3 Bias} [\textit{reverse-coded}]: derogatory attitude towards people who hold opposing views
    \item \textbf{F4 Openness} [\textit{reverse-coded}]: beliefs that some topics are not up for discussion
\end{enumerate}
To translate the scale to the fast-paced MTurk platform and to our conversational receptiveness task, we reduce the quantity and transform the statements into two questions per factor (Appendix Table \ref{tab:receptivness_factors}). The questions compare the receptiveness of the reframed or baseline reply to the original reply, though annotators are not told which is which. We use a 7-point Likert scale for each question: -3 indicates the original reply is much more receptive, +3 indicates the opposite. Answers are averaged over the eight questions to determine our receptiveness index for each example.

\subsection{Results}
\label{sec:results}
To compare mean receptiveness scores between our reframed and baseline replies, we use a mixed effects logistic regression model via the {\tt lme4} package \cite{bates_mixed_effects}. We model the receptiveness score with a random effect for the comment ID and fixed effect for the strategy used.

We show the regression co-efficients in Appendix Table \ref{tab:statistics_regression_coefficients}. The estimated marginal mean and the 95\% CI receptiveness scores of the strategies and the baseline  are shown in {Figure \ref{fig:receptiveness_scores}}. A marginal mean of zero implies the reframed reply is no different from the original reply with respect to receptiveness. From {Figure \ref{fig:receptiveness_scores}, we find that neither the strategies nor baselines overlap with 0, indicating that they are more receptive than the original. The reframes using our strategies have a statistically significant higher receptiveness index than both baselines. This indicates that annotators consider generations from our strategies more receptive than the paraphrases or baseline receptive reframes. In other words, the replies generated using our social science-informed framework are more aligned with human notions of receptiveness, compared to a generic prompting approach. The mean differences, test statistics and p-values are in Appendix Table \ref{tab:statistics_receptiveness_contrasts}.

\begin{figure}[!t]
    \centering
    \includegraphics[width=\linewidth]{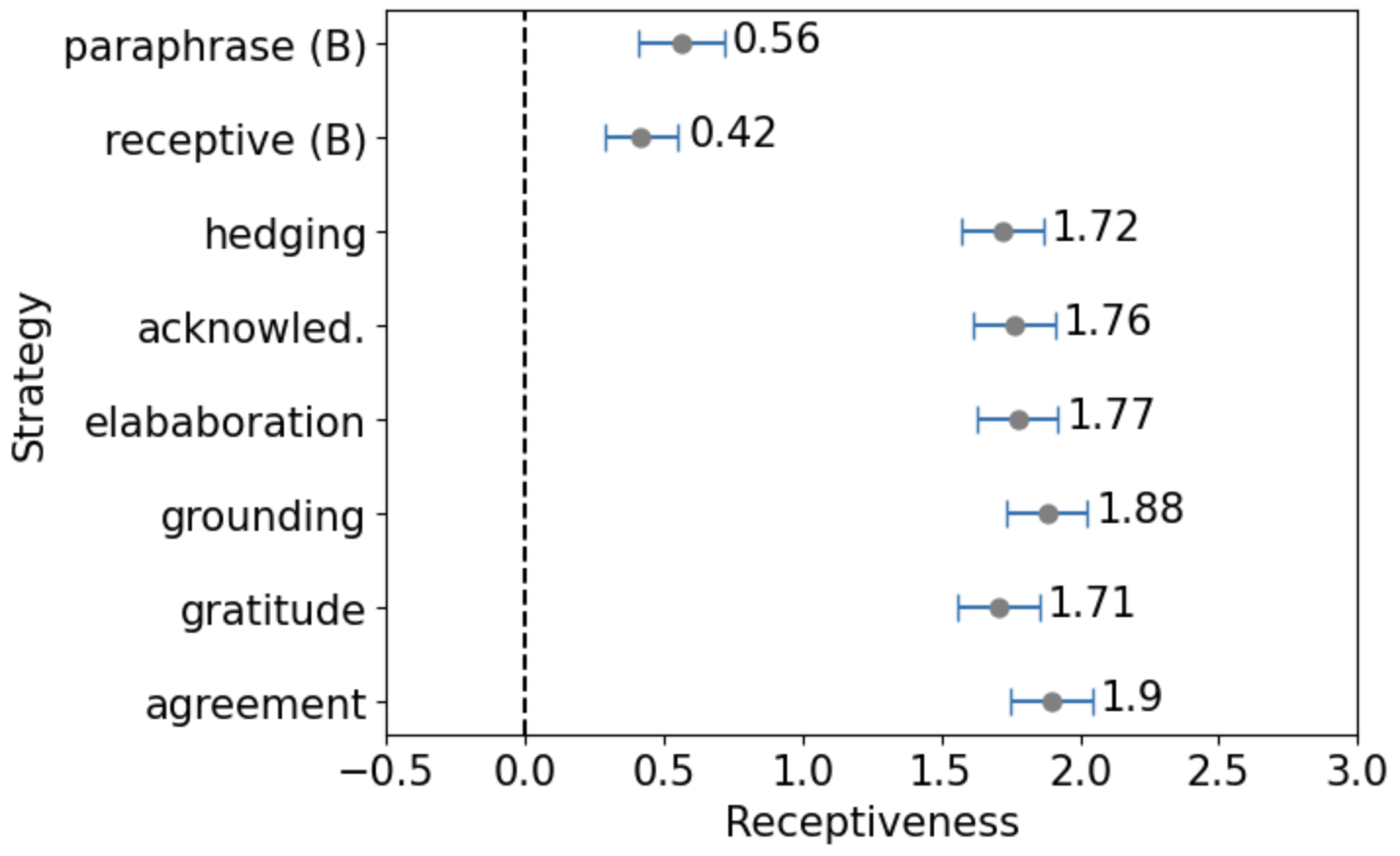}
    \caption{Receptiveness scores for reframes compared to the 
    original replies. The dot indicates the mean. Lines show the 95\% CI. Positive values indicate that the reframed reply is more receptive than the original. ``B'' indicates a baseline.}
    \label{fig:receptiveness_scores}
\end{figure}

\begin{table}[!t]
    \small
    \centering
    \begin{tabular}{l|rrrrr}
\toprule
       {\bf Strategy}  & \multicolumn{5}{c}{\bf Receptiveness Score} \\ & {\bf F1} &   {\bf F2} &   {\bf F3} &   {\bf F4} &  {\bf \!\!\!Avg.}\\
\midrule
        Hedging  & {\bf 1.79} & 1.59 & 1.02 & 1.75 &   1.54\\
Acknowledge  & 1.72 & 1.66 & 1.06 & {\bf 1.79} &   1.56\\
    Elaboration  & {\bf 1.85} & 1.71 & 1.09 & 1.78 &   1.61\\
      Grounding  & 1.86 & 1.72 & 1.11 & {\bf 1.90} &   1.65\\
      Gratitude  & {\bf1.81} & 1.65 & 1.00 & 1.71 &   1.54\\
      Agreement  & {\bf 1.98} & { 1.81} & { 1.17} & { 1.89} &   { 1.71}\\
      \midrule
      Avg. & {\bf 1.84} & 1.69 & 1.07 & 1.80 & {1.60}\\
\bottomrule
\end{tabular}
    \caption{Receptiveness over factors: Emotion (F1), Curiosity (F2), Bias (F3), and Openness (F4).  The highest value in each row is bolded, indicating the factor that contributes most towards a particular strategy.}
    \label{tab:receptiveness_factor}
    \vspace{-1em}
\end{table}
\subsection{Factor breakdown of receptiveness}
We have seen that our framework-generated reframes are more aligned with human perceptions of receptiveness, but why is this the case? We investigate this by examining the  receptiveness scores split by the individual factors. To do so, we averaged the values for the two questions in each factor. We can see from Table \ref{tab:receptiveness_factor} that while our reframes appear to decrease negative emotions, increase openness, and increase curiosity towards opposing views, they interestingly do \textit{not} do as well in reducing bias towards the opposing party. We discuss this further in Section \ref{sec:discussion}. Such a factor analysis (drawing upon social science theory) provides deeper insight into the reasoning for the receptiveness scores than would be possible with the direct receptive baseline. 

\subsection{Toxicity effects on perceived reception}
\label{sec:context_toxicity}

In this section, we analyze whether reframing may improve receptiveness at different rates based on the toxicity of the original reply. Toxicity scores $\in[0,1]$ are predicted using the Perspective API. 

Prior studies have validated use of this API for toxicity classification on Reddit \cite{rajadesingan2020quick}. We use the mixed effects regression model as before but also include an interaction term for toxicity and strategy. As our previous regression did not reveal differences in receptiveness between strategies, we only compare differences between use of any strategy to the baselines. 

We show the results of the regression in {Figure \ref{fig:toxicity_receptiveness}}. The receptiveness scores are split by toxicity for receptiveness reframes (all strategies together) and each baseline. We operationalize low toxicity as [0, 0.5), medium as [0.5, 0.7), and high as [0.7, 0.9]. As noted in Section \ref{sec:task}, we exclude examples that have a toxicity score above 0.9. 

We find that the difference in perceived receptiveness of our reframes based on the toxicity level of the original reply is statistically significant (using the Ward test) for some toxicity levels; the difference between low and high toxicity and between low and medium is significant, but the difference between medium and high toxicity is not. The differences between perceived receptiveness for the two baselines are not significant between any of the toxicity levels of the original replies. The mean differences between toxicity levels, test statistics, and p-values for both receptiveness reframes and baselines are shown in Appendix Table \ref{tab:statistics_toxicity_contrasts}. 

These results suggest that our receptive reframes have higher impact on the more toxic comments. Neither paraphrase nor receptive baseline generations were perceived to be any more receptive on toxic comments, further drawing the distinction between our reframing strategies and the baselines.

\begin{figure}[!t]
    \centering
    \includegraphics[width=\linewidth]{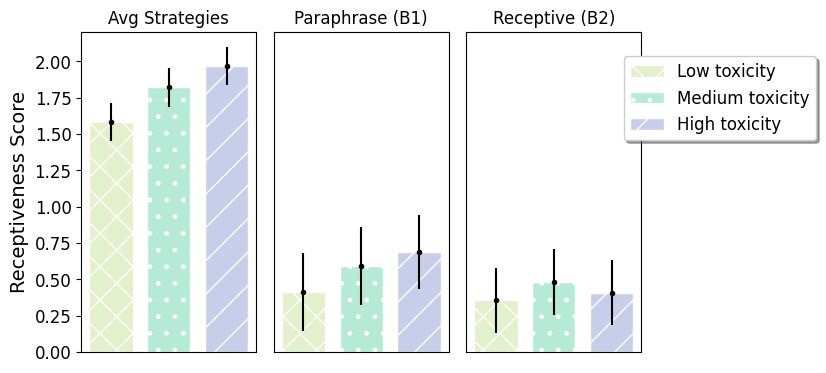}
    \caption{Receptiveness at different toxicity levels (per the PerspectiveAPI): low [0, 0.5), medium [0.5, 0.7), and high  [0.7, 0.9]. Results are shown for baselines B1 and B2 and the average over receptive-reframed replies. }
    \label{fig:toxicity_receptiveness}
    \vspace{-1em}
\end{figure}

\section{Discussion}
\label{sec:discussion}
\vspace{-0.5em}

\paragraph{Differences in receptiveness between factors}
As discussed in Section \ref{sec:receptiveness_experiment}, our reframes have a smaller effect on reducing bias (F3) than on any of the other factors. This presents promising opportunities for future work on receptiveness reframing aimed at generating reframes that reduce outgroup bias. For example, this could include explicit instruction to reduce bias in addition to our prompts, or attempting to signal \textit{impartiality} or \textit{humility} \cite{alsheddi2020humility, fisher2000intergroup} as ways to implicitly reduce bias. However, these methods might result in a tradeoff between reducing bias and meaning preservation. See  {\em Limitations} for more discussion.

\paragraph{Combining multiple strategies}
While we only generate reframes using one strategy at a time, future work could explore combining  strategies. We do find that certain receptiveness reframes contain strategies that were not explicitly prompted for; this was especially prominent with hedges, which were commonly added to generations using other strategies. Understanding how different combinations of strategies might affect perceived receptiveness would allow us to refine the current framework. Future work could also automatically optimize for which strategies to use in different contexts. 

\paragraph{Receptiveness for polarization} 
Increasing affective polarization has contributed to partisan gridlock \cite{hetherington2020washington}, and more recently, limited compliance with COVID-19 mitigation policies \cite{druckman2021affective}. A major intervention known to reduce polarization is actually fairly straightforward: to engage in conversations with the other side \cite{levendusky2021we}. We believe that our computational framework for reframing, if integrated into conversation spaces, could improve the quality of conversations between opposing partisans, potentially reducing polarization \cite{Argyle2023-mz}. 

\paragraph{Receptiveness for content moderation at scale}
The ability to automatically reframe content could strengthen moderation at scale. Moderators could suggest receptiveness-signaling reframes to users as rewrites for comments which might be removed otherwise, discouraging users from commenting further \citep{Myers_West2018-ew}. 
Reframing one's own comments also provides an opportunity to learn from mistakes, which is known to lead to greater improvement in skill learning \citep{mason_yerushalmi_2016}.

\paragraph{Beyond receptiveness}
The framework described in this paper could be easily applied to address factors beyond receptiveness, which is not the only deliberative ideal that helps promote constructive deliberation. To encourage quality discussion across disagreements, deliberation scholars have identified multiple deliberative ideals in prior work, such as rationality (reason-giving) and common good orientation \cite{Bachtiger2018-hj}. While our receptiveness reframing aims to modify the tone of the discussion while preserving the meaning, one could use the same framework of incorporating low-level linguistic strategies from social science theory, but focus on improving the substance of the discussion. For example, to encourage more rational discussions, one could rewrite comments to both ask and make (or nudge users to make) more justification of claims made. Similarly, one could aim to rewrite comments to make arguments with a common good orientation rather than ones with narrow group interests.

\section{Conclusion}
\label{sec:conclusion}
\vspace{-0.5em}

This work illustrates how transforming the social science construct of receptiveness into a computational framework can make LLM generations more aligned with human perceptions. We found that our reframed replies are perceived to be better aligned with human notions of receptiveness than a reply generated with a generic receptiveness prompt. We also analyzed how our reframes are perceived in terms of the different receptiveness factors, and how some (emotion, curiosity, openness) are perceived to have improved more than others (bias). Finally, we demonstrated that our reframes appear to be more effective when the original replies are more toxic, and discuss applications of receptiveness-signaling for content moderation and reducing polarization. 

\section*{Limitations}
\label{sec:limitations}

\paragraph{Overlap between strategies}
We curate six linguistic strategies from social science literature that signal receptiveness to another person. There is overlap between some of these strategies, particularly between agreement, acknowledgement, and grounding. As discussed in Section \ref{sec:validation}, the overlap between these strategies was along expected lines. It appears that the agreement prompt does not always reframe replies with explicit agreement and instead appears to simply acknowledge the speaker's perspective. This is likely because we prompt the model to make ``explicit any agreement with the comment.'' and in the absence of agreement expressed in the original reply, the model appears to default to acknowledge rather than to agree. For a more distinct strategy, we might instead study \textit{token agreement}, where we don't try to make any agreement explicit (because there usually isn't any) and instead simply reframe in the form ``yes, but...''.

\paragraph{Reframing and meaning preservation tradeoff}
We acknowledge that there exists a tradeoff between receptive reframing and meaning preservation; the most effective receptive reframe might not preserve meaning to the same degree as a more subtle reframe, and conversely, optimizing for meaning preservation alone would likely lead to reframes that are not as receptive.

\paragraph{Confounds beyond linguistic strategies}
The model-generated reframes make some changes beyond purely adding phrases that signal a particular receptiveness strategy. Model generations tend to fix grammatical and other syntax errors, make language more formal, and otherwise slightly change the meaning of the original replies. We attempt to account for changes in syntax and formality by comparing to the other model-generated baselines, but we do not account for other potential changes in meaning, besides what we measure (imperfectly) with automatic metrics in Section \ref{sec:validation_meaning}.

\paragraph{Lack of comparison between strategies} While we would have liked to evaluate the effectiveness between pairs of strategies, the size of our dataset does not allow for meaningful tests of significance for multiple pairwise comparisons and interaction analyses. While we illustrate how all strategies show strong improvements over the baselines, further between-strategy analysis would be valuable.

\paragraph{30 word limit to comments and replies} As we mention in Section \ref{sec:task}, 30-word comments and replies are around the average word length of all comments and replies in our dataset. However, we note that the effect of reframing on longer messages may be different.

\paragraph{Receptiveness measure is US-centric}
Our study is very US-centric; we use subreddits whose topics are mostly centered in the US (r/democrats, r/Republican), and therefore only selected annotators who live in the US. As a result, our measure of receptiveness and our results are US-centric.

\paragraph{Limitations of the Perspective API}
We rely on the toxicity classifier from the Perspective API for our toxicity classifications, but this is based on only one notion of toxicity and may not account for others. In addition, classifiers aren't perfect and it is likely there are some errors in the toxicity classifications.

\section*{Ethical Considerations}

\paragraph{Exposing annotators to toxic content}
Our data contains some toxic content, which is shown to annotators during both our reasonability evaluation task and receptiveness experiment. We note that this exposure can be harmful to annotators \cite{shmueli-etal-2021-beyond}.

\paragraph{Reframes might still be offensive}
We acknowledge that our generated reframes might still be offensive, as we seek to preserve meaning with the original comments. We do not intend potentially offensive reframes to be deemed non-toxic simply because they incorporate one of our strategies.

\paragraph{Reframes might be biased}
We note that model generations might be biased towards particular social groups and/or political identities \cite{santurkar_2023}, and this could affect the content of our receptiveness reframes.

\paragraph{Intended use of data}
Our reframes and annotations are released for research use. We emphasize that (as discussed above) reframes might still be offensive or biased, and this should be noted when this data is being used.

\section*{Acknowledgements}

We thank the online workers who provided annotations for this study, and the anonymous reviewers for their valuable feedback. This research was supported in part by the Knight Foundation and by Good Systems\footnote{\url{http://goodsystems.utexas.edu/}}, a UT Austin Grand Challenge to develop responsible AI technologies. We acknowledge the Texas Advanced Computing Center (TACC) for compute resources. The statements made herein are solely the opinions of the authors and do not reflect the views of the sponsoring agencies.

\bibliography{anthology,custom}

\appendix

\section{Ensuring quality annotation}\label{appendix:quality_annotation}
To ensure quality annotation, we apply two strategies to curate our annotator pool: (1) sourcing workers already pre-qualified by Amazon for its Ground Truth workforce\footnote{\scriptsize \url{https://aws.amazon.com/sagemaker/groundtruth/}}, and (2) filtering other workers ourselves, requiring at least 97\% accuracy and 5000 completed hits. We also require workers to pass a qualification test consisting of four examples similar to our main task. Finally, given the US-centric nature of our data, we require workers to be US-based. We pay workers \$10/hr. 

\section{Prompts and Examples for Model Reframes}
\label{appendix:prompts_and_examples}

\subsection{Prompts}
\label{appendix:prompts}

Pilot experiments led to the following prompt template across strategies: ``{\tt Rewrite the reply using <strategy>, following the examples. <Strategy> means <strategy definition>. Use at most 30 words.}'' The final strategy definitions used in the template were: 
\begin{description}[topsep=0.5em,itemsep=-0.5em]
{\item[Hedging] ``adding words or phrases that soften the force of a statement''
\item[Acknowledgement] ``conveying understanding and validating a person’s feelings''
\item[Elaboration] ``asking for further details, clarifying understanding, or repeating what was heard''
\item[Grounding] ``establishing common ground, and explicitly mentioning mutual knowledge, beliefs, or assumptions''
\item[Gratitude] ``showing appreciation for someone’s thoughts and opinions''
\item[Agreement] ``making explicit any agreement with the comment''}
\end{description}
For the last strategy, we ultimately used ``compromise and agreement'' for {\tt <strategy>}, instead of merely ``agreement'', since the latter tended to cause the model to simply agree with the comment (even if this contradicted the original reply); adding ``compromise'' helped to remedy this. Each prompt is followed by the five in-context examples in \ref{appendix:examples}.

Our zero-shot paraphrase baseline was: ``Paraphrase the following reply to the comment. Use at most 30 words''. For the receptiveness baselines, we use the prompt: ``Rewrite the following reply to the comment to be more receptive. Use at most 30 words.''

\subsection{Examples}
\label{appendix:examples}
\begin{enumerate}
    \item \textbf{Comment:} Worth it to regain influence over the top executive who writes laws that affect one's life.\newline
    \textbf{Reply:} You think you have any influence over BoJo, Cummings, Rees-Mogg, Raab? lol. They wouldn't care if you died in a tower block fire.\newline
        \textbf{Hedging:} I’m not sure you would really have any influence over BoJo, Cummings, Rees-Mogg, and Raab. They likely wouldn’t even care if you died in a tower block fire.\newline
        \textbf{Acknowledgement:} I understand you think you have influence over BoJo, Cummings, Rees-Mogg, Raab, but I don’t think they would care if you died in a tower block fire.\newline
        \textbf{Elaboration:} From what I understand, you’re saying you would have influence over BoJo, Cummings, Rees-Mogg, Raab. However, they likely wouldn’t care if you died in a tower block fire.\newline
        \textbf{Grounding:} I agree it would be nice to have influence overBoJo, Cummings, Rees-Mogg, and Raab, but I don’t think it would actually happen. They wouldn't care if you died in a tower block fire.\newline
        \textbf{Gratitude:} I appreciate your perspective, but I don’t think you have any influence over BoJo, Cummings, Rees-Mogg, and Raab. They probably wouldn't care if you died in a tower block fire.\newline
        \textbf{Agreement:} I agree it would be nice to regain influence over top executives like BoJo, Cummings, Rees-Mogg, Raab. But I don’t think you actually would, they probably wouldn't care if you died in a tower block fire.
    \item \textbf{Comment:} I love all of it. Except the proposed gun legislation. That might actually make me walk away from Biden. It's too much and too impractical. Which is a shame because I was very excited and ready to vote for him. \newline 
    \textbf{Reply:} Don't be a single issue voter, it's an absolutely ridiculous position to hold.\newline
        \textbf{Hedging:} Perhaps being a single issue voter is not the best position to hold.\newline
        \textbf{Acknowledgement:} I would feel the same way if gun legislation was an important issue to me, but it’s important to not be a single issue voter.\newline
        \textbf{Elaboration:} I think you’re saying that the one issue of gun legislation is turning you against Biden. I don’t think being a single issue voter is the best position to hold.\newline
        \textbf{Grounding:} I also think certain issues are more important than others, but I don’t think being a single issue voter is the best position to hold.\newline
        \textbf{Gratitude:} Thank you for sharing your view, but I don’t think being a single issue voter is the best position to hold.\newline
        \textbf{Agreement:} I agree that gun legislation is an important issue, but I don’t think being a single issue voter is the best position to hold.
    \item \textbf{Comment:} "Um akshywally, cops kill more whites tham blacks, so while they're still murderers and thugs, at least they're not racist." OP is an idiot. \newline
    \textbf{Reply:} Cops aren't murderers and thugs. Like 1\% of them are bad. You need to respect the people that risk their lives to protect innocent people. The ones that don't do that are the bad 1\%.\newline
        \textbf{Hedging:} I don’t think all cops are murderers and thugs. Maybe around 1\% of them are bad. It might be good to respect the people that risk their lives to protect innocent people. Admittedly, the cops that don’t protect people are bad 1\%.\newline
        \textbf{Acknowledgement:} I realize this is a complex situation, but cops aren’t all murderers and thugs, maybe like 1\% of them are bad. I see how they can all seem bad, but it’s important to respect the people that risk their lives to protect innocent people. The ones that don’t do that are the bad 1\%.\newline
        \textbf{Elaboration:} Would you mind clarifying what you mean about cops being murderers and thugs? I don’t think all cops are murderers and thugs. Like 1\% of them are bad. You should respect the people that risk their lives to protect innocent people. The ones that don't do that are the bad 1\%.\newline
        \textbf{Grounding:} I also believe that cops that don’t risk their lives to protect innocent people are bad, but I think that’s only like 1\% of them. Cops aren’t all murderers and thugs.\newline
        \textbf{Gratitude:} I appreciate your opinion, but I don’t think all cops are murderers and thugs. Like 1\% of them are bad. It’s important to respect the people that risk their lives to protect innocent people. The ones that don't do that are the bad 1\%.\newline
        \textbf{Agreement:} I think you’re right that some cops kill people, but they aren’t all murderers and thugs. Like 1\% of them are bad. It’s important to respect the people that risk their lives to protect innocent people. The ones that don’t help people are the bad 1\%.
    \item \textbf{Comment:} Is this a homework assignment? Either way, easy enough to learn about what white privilege is if you just do a little Google searching. \newline
    \textbf{Reply:} Nope I want people to tell me what privileges they have over other races. Not some dumb bs like "the black male makes .8 cent of the white males dollar"\newline
        \textbf{Hedging:} Actually, I think I want people to tell me what privileges white people have over other races. Statistics like "the black male makes .8 cent of the white males dollar" usually aren’t very meaningful to me.\newline
        \textbf{Acknowledgement:} I know you could easily learn about white privilege online, but I want people to tell me what privileges they have over other races. While I understand that statistics like "the black male makes .8 cent of the white males dollar" are important, they aren’t as meaningful to me.\newline
        \textbf{Elaboration:} I think you’re saying that it’s easy to learn about white privilege from searching Google, but I want people to tell me what privileges they have over other races. Not just statistics like "the black male makes .8 cent of the white males dollar".\newline
        \textbf{Grounding:} I also think it’s good to learn about white privilege, but I want people to tell me what privileges they have over other races, rather than finding statistics like "the black male makes .8 cent of the white males dollar" online.\newline
        \textbf{Gratitude:} I appreciate your comment that we can learn about white privilege online, but I want people to tell me what privileges they have over other races, not just statistics like "the black male makes .8 cent of the white males dollar". \newline
        \textbf{Agreement:} It makes sense that you can learn about white privilege by just Googling, but I want people to tell me what privileges they have over other races, not just statistics like "the black male makes .8 cent of the white males dollar".
    \item \textbf{Comment:} You dont scare people when the recovered total is almost as big as the total number. \newline
    \textbf{Reply:} Dude we are having a 9/11 in deaths from corona everyday and it has become the leading cause of death in the u.s.. that still sounds like a pretty big problem \newline
        \textbf{Hedging:} It seems like we are actually having a 9/11 in deaths from corona everyday and it has become the leading cause of death in the u.s.. That still sounds like a relatively big problem.\newline
        \textbf{Acknowledgement:} I agree that for a large number of people the corona virus is recoverable. However, we are having a 9/11 in deaths from corona everyday and it has become the leading cause of death in the u.s., which still sounds like a pretty big problem.\newline
        \textbf{Elaboration:} Are you saying that because so many people recover, it means the corona virus isn’t a problem? We are actually having a 9/11 in deaths from corona everyday and it has become the leading cause of death in the u.s.. that still sounds like a pretty big problem.\newline
        \textbf{Grounding:} I also believe that it’s good that many people recover from the coronavirus, but we are having a 9/11 in deaths from corona everyday and it has become the leading cause of death in the u.s.. that still sounds like a pretty big problem.\newline
        \textbf{Gratitude:} I appreciate you bringing up that many people do recover from corona, but we are still having a 9/11 in deaths from the disease everyday and it has become the leading cause of death in the u.s.. that still sounds like a pretty big problem.\newline
        \textbf{Agreement:} I agree that a lot of people recover from covid, but we are having a 9/11 in deaths from corona everyday and it has become the leading cause of death in the u.s.. that still sounds like a pretty big problem.
\end{enumerate}

\section{Additional Validation Results}
\label{appendix: validation_results}
We study differences in trigrams found in generated reframes compared to the original replies. The overlap of these trigrams between the different strategies is given in Figure \ref{fig:overlap}. Overlap between a pair of reframes is defined as the intersection of their trigrams divided by the union of their trigrams.

\begin{figure}[!t]
    \centering
    \includegraphics[width=\linewidth]{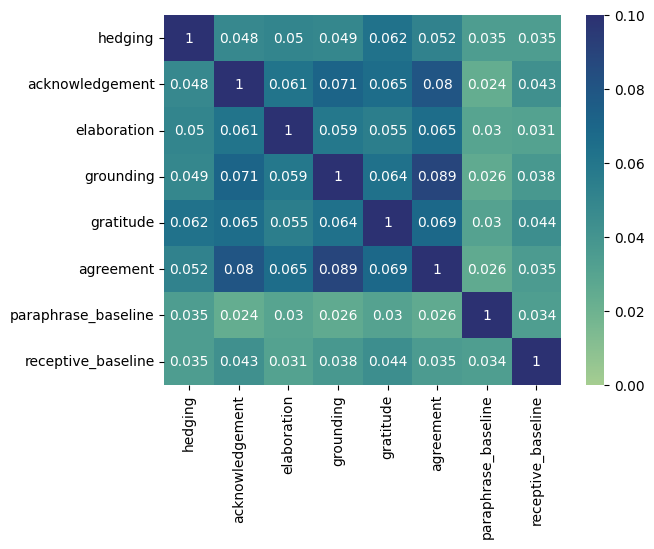}
    \caption{Pair-wise overlap between each of our strategy-reframes and baseline reframes. Overlap was calculated as the intersection of trigrams over the union, for each pair of reframes.}
    \label{fig:overlap}
\end{figure}

\section{Additional Qualitative Examples}
\label{appendix: qualitative examples}
Examples of receptiveness reframes for comment-reply pairs are listed in Table \ref{tab:additional_qualitative_reframes}. Examples of predicted contradictions between original and reframed replies are shown in Table \ref{tab:contradiction_examples}. Examples of original replies that might be seen as less \textit{reasonable} than the model-generated reframed reply are shown in Table \ref{tab:original-replies-bad}. Examples of comments and replies for varying levels of reply toxicity are shown in Table \ref{tab:toxicity_examples}.

\begin{table*}[!t]
    \footnotesize
    \centering
    \begin{tabular}{ p{0.085\textwidth} p{0.26\textwidth} p{0.26\textwidth} p{0.26\textwidth}}
    \toprule
    & Example 1 & Example 2 & Example 3\\
    \midrule
    Comment & Am I the only person reasonable enough to believe that both Democrats and Republicans love this country and that they both disagree on the ways that make it lovely? & I'm white and my family worked in the fields of Alabama. Lived in a tiny house with no running water until the 70's. Wtf & No. Stop bringing racism into a movement trying to get rid of it.\\
    Reply & Nice false choice fallacy. Democrats hate America. Democrats hate Trump, the people who voted for Trump and they want retribution. End of story lol & You're white. You and your entire family are automatically privileged. & Oh i said black people good and whites bad am not racist am "protesting"\\
    Hedge rewrite & Some people tend to generalize and assume that Democrats hate America, but perhaps it is possible that both Democrats and Republicans disagree on how to make America great. & While I understand that you and your family have faced significant challenges, it is still important to acknowledge the existence of white privilege in society. & It might not be helpful to bring racism into a movement trying to get rid of it. Saying things like "black people good and whites bad" could be seen as divisive, even if it’s meant to be a form of protesting.\\
    Ack rewrite & I recognize that both Democrats and Republicans love this country, but it seems like the polarizing rhetoric from both sides can make it difficult to see that they ultimately want the same thing for America. & I understand that assumptions about privilege can be hurtful and dismissive. It is important to recognize that privilege is not a one-size-fits-all experience and your family's struggles are valid. & I understand your frustration, but it's important to recognize that the Black Lives Matter movement is about addressing systemic racism and promoting equality, not about demonizing any particular race.\\
    Elab rewrite & I’m hearing your opinion that Democrats and Republicans both love this country but disagree on how to make it better. However, I don't think it’s fair to say that Democrats hate America and want retribution against Trump and his supporters. & It sounds like you're saying that just because you're white, you're automatically privileged. I don't agree with that. Your family had a tough life living in a tiny house with no running water until the 70s in Alabama. & I think you’re saying that you don’t want racism brought into a movement trying to get rid of it. I believe it's important to focus on the issue of racism and not make generalisations, like "black people good and whites bad".\\
    Ground rewrite & I agree that it’s possible for both Democrats and Republicans to love this country and simply disagree on how to make it better. However, it’s unfair to make a false choice fallacy by suggesting that Democrats hate America and want retribution against those who voted for Trump. & I understand that not all white people have had easy lives or privileges, but I think it’s important to acknowledge that there is still systemic racism that affects people of color. & I agree that bringing racism into a movement trying to get rid of it is not helpful, but I don’t think it’s fair to assume that all protesters are saying "black people good and whites bad."\\
    Grat rewrite & Thank you for your comment, but I don't believe it's a fair assumption to say that Democrats hate America. Both parties love this country, they just have different ways of approaching and achieving their goals. & Thank you for sharing your personal experience. While privilege certainly exists, it's important to recognize that every individual's situation is unique and not defined solely by their race. & I hear your point about not wanting to bring racism into a movement trying to eliminate it, but making sweeping statements about any race being inherently good or bad can be divisive and counterproductive.\\
    Agree rewrite & I agree that both Democrats and Republicans love this country, but it's unfair to say that Democrats hate America. They may just have different visions for what makes it lovely. & I acknowledge that your family faced struggles and challenges. However, it's important to recognize that white privilege exists in society, regardless of individual circumstances. & I understand that it's not helpful to bring racism into a movement trying to eliminate it. Let's focus on constructive ways to address racism without demonizing any particular group.\\
    \bottomrule
    \end{tabular}
    \caption{Additional qualitative examples of Reddit comments and replies, as well as the reframes generated for each strategy.}
    \label{tab:additional_qualitative_reframes}
\end{table*}

\begin{table*}[!t]
    \small
    \centering
    \begin{tabular}{ p{0.4\textwidth} p{0.5\textwidth} }
    \toprule
    Original Reply & Reframed Reply \\
    \midrule
    Nice false choice fallacy. Democrats hate America. Democrats hate Trump, the people who voted for Trump and they want retribution. End of story lol & I agree that both Democrats and Republicans love this country, and it's important to recognize that they have different perspectives on what policies will make it better. \newline\\
    Dude we hold everyone to higher standards then a kid. A kid curses and everyone says it's wrong. An adult curses and no one cares??? \newline& I understand your point, but we should hold everyone, including the president, to a higher standard than kids. Cursing is wrong regardless of age. \newline\\
    Dude we hold everyone to higher standards then a kid. A kid curses and everyone says it's wrong. An adult curses and no one cares??? \newline& You're right that holding kids to a higher standard than the president is problematic. We should hold everyone, including the president, to a higher standard. \newline \\
    So if they were black it would have been fine you racist and the child where not harmed & I share your horror at the thought of children being handcuffed. Regardless of race, it's unacceptable. I'm relieved to hear the child wasn't harmed. \newline\\
    Just playing by the ``ballot harvesting'' rules that the Democrats put in place. & While I understand the Republicans are using ``ballot harvesting'' rules, it's concerning to hear about the use of ``unofficial'' ballot boxes and potential voter fraud. \\
    \bottomrule
    \end{tabular}
    \caption{Examples of contradictions (by the out-of-the-box NLI model) between original replies and model-generated reframes.}
    \label{tab:contradiction_examples}
\end{table*}

\begin{table*}[!t]
    \small
    \centering
    \begin{tabular}{ p{0.3\textwidth} p{0.3\textwidth} p{0.3\textwidth}}
        \toprule
        {\bf Comment} & {\bf Reply} & {\bf Reframe} \\
        \midrule
        ``{Well creepy isnt just a nickname anymore. It is an earned title. And let the games begin.}'' & ``{Like trump hasn't done the same shit? Tunnel vision at work}'' & ``{Are you saying that Trump hasn't done things that could also be considered creepy? It sounds like you might not be considering everything.}'' \\
        ``{whales are not fish any more than seals or otters are. is that what u meant?}'' & ``{I am positive that is not what the artist meant. Saving anything doesn't mean fuck everything else.}'' & ``{I understand your point about whales not being fish. But I'm sure the artist intended to convey saving wildlife in general, not disregarding other species.}'' \\
        ``{No, this is going to be the start of the end of the republican party.}'' & ``{This idiot is a genius in campaigning. I am dooming pretty hard for 22 and 24.}'' & ``{I also think this could be the start of the end for the Republican party, but I'm worried about '22 and '24 since this "idiot" is actually a genius in campaigning.}'' \\
        \bottomrule
    \end{tabular}
    \caption{Examples of human-written original replies that might not make direct sense from the original comment, compared to generated reframes that take into account context from the comment.}
    \label{tab:original-replies-bad}
\end{table*}

\begin{table*}[]
    \small
    \centering
    \begin{tabular}{ p{0.3\textwidth} p{0.3\textwidth} l}
        \toprule
        {\bf Comment} & {\bf Reply} & {\bf Reply toxicity} \\
        \midrule
        ``"Education funding from property taxes". How else do they expect free public schools?'' & ``I don't think property taxes in a certain area directly benefit that neighborhood's schools. It just all goes into the general fund, as far as I know.'' & 0.006 \\
        ``The GOP sucks but are we not going to push back against representative Waters' comments? She should not have said what she said... "we need to be more confrontational"'' & ``What is wrong with that? She understands the people. If there was no murder Minnesota be burning right now'' & 0.251 \\
        ``How do all of these people get shot when it's almost impossible to own a gun in NYC'' & ``It's really not that hard to get a gun in New York... Maybe a long gun but you can kill people with a pistol.'' & 0.508 \\
        ``We may disagree on stuff, but the left doesn't "hate America.'' & ``Bullshit they don't. Oh, sure. They don't hate the land. They just hate the founders, our culture, everything we stand for and our entire way of life.'' & 0.751 \\
        ``HTF did anyone think he make a great President? He was a P.O.S. all his life..'' & ``Mindless idiots that are too fucking stupid to know that they're fucking mindless idiots, that's who.'' & 0.968 \\
        \bottomrule
    \end{tabular}
    \caption{Examples of Debagreement comments and replies for varying levels of reply toxicity.}
    \label{tab:toxicity_examples}
\end{table*}

\section{Receptiveness Factors}
\label{appendix:factors}
Table \ref{tab:receptivness_factors} contains each of the four factors that make up the receptiveness index, as well as the corresponding questions that were given to annotators to determine factor scores.

\begin{table*}[!t]
    \small
    \centering
    \begin{tabular}{ p{0.4\textwidth} p{0.4\textwidth} }
    \toprule
    {\bf Receptiveness Factor} & {\bf Questions}\\
    \midrule
    {(1)} Negative emotional reactions towards opposing views& (a) \negative{Which reply would be more likely to make you angry when you read it?}\\
 & (b) \positive{Which reply would be less likely to make you mad?}\\
 \midrule
    {(2)} Curiosity towards opposing views& (a) \negative{Which reply makes you feel less interested in finding out why the user has a different view than you?}\\
 &(b) \positive{Which reply makes you feel genuinely curious to find out more about why they have a different opinion than you do?}\\
 \midrule
    {(3)} Bias towards people holding opposing views& (a) \negative{Which reply makes you feel like the user's view is biased by what would be best for them and their group?}\\
 &(b) \negative{Which reply makes you feel like the user's opinion is based on preconceived notions of what would be best for them and their group?}\\
 \midrule
    {(4)} Beliefs that some topics are off limits& (a) \negative{Which reply makes you feel like the user is unwilling to further discuss the issue?}\\
 &(b) \negative{Which reply makes you feel like the issue is just not up for debate?}\\
 \bottomrule
    \end{tabular}
    \caption{Receptiveness factors and corresponding questions asked to users. Orange indicates the questions are reverse coded.}
    \label{tab:receptivness_factors}
\end{table*}

\section{Statistics Tests}
\label{appendix:statistics_tests}
This section contains the details of the statistical tests described in the paper. Table \ref{tab:t_test_statistic_plausibility} contains the mean differences, test statistics, and p-values for the t-test on results from the reasonability annotation task. Table \ref{tab:statistics_regression_coefficients} contains the full regression coefficients and test statistics from the receptiveness annotation task. Table \ref{tab:statistics_receptiveness_contrasts} contains the mean differences, test statistics, and p-values for the Wald test on results from the receptiveness annotation task. And Table \ref{tab:statistics_toxicity_contrasts} contains the mean differences, test statistics, and p-values for the Wald test on results from the toxicity interaction analysis.

\begin{table}[!t]
    \centering
    \begin{tabular}{cccc}
    \toprule
    \textbf{Strategy} & \textbf{Mean diff} & \textbf{Test stat.} & \textbf{p-value} \\
    \midrule
    Hedging & 0.454 & 3.814 & < 0.0001\\
    Acknowled. & 1.008 & 7.776 & < 0.0001\\
    Elaboration & 0.724 & 5.676 & < 0.0001\\
    Grounding & 0.917 & 7.360 & < 0.0001\\
    Gratitude & 0.729 & 5.113 & < 0.0001\\
    Agreement & 0.852 & 6.384 & < 0.0001\\
    \bottomrule
    \end{tabular}
    \caption{Mean difference between each strategy and the original replies, paired t-test statistics and p-values for the reasonability annotation task. The paired t-test is conducted for each strategy compared to the original reply.}
    \label{tab:t_test_statistic_plausibility}
\end{table}

\begin{table}[!t]
    \centering
    \begin{tabular}{cccc}
    \toprule
    \textbf{Strategy} & \textbf{Coefficient} & \textbf{Test stat.} & \textbf{p-value}\\
    \midrule
    Paraphrase (B) & 0.565 & 7.132 & < 0.0001\\
    Receptive (B) & 0.416 & 6.085 &  < 0.0001\\
    \midrule
    Hedging & 1.153 & 11.479 & < 0.0001\\
    Acknowled. & 1.196 & 12.048 & < 0.0001\\
    Elaboration & 1.209 & 12.138 & < 0.0001\\
    Grounding & 1.316 & 13.345 & < 0.0001\\
    Gratitude & 1.142 & 11.463 & < 0.0001\\
    Agreement & 1.153 & 13.318 & < 0.0001\\
    \bottomrule
    \end{tabular}
    \caption{Regression coefficients for the mixed effects logistic regression model.``(B)'' indicates a baseline.}
    \label{tab:statistics_regression_coefficients}
\end{table}

\begin{table}[!t]
    \centering
    \begin{tabular}{cccc}
    \toprule
    \textbf{Strategy} & \textbf{Mean diff} & \textbf{Test stat.} & \textbf{p-value} \\
    \midrule
    Hedging & 1.15 & 11.478 & < 0.0001\\
    Acknowled. & 1.20 & 12.048 & < 0.0001\\
    Elaboration & 1.21 & 12.137 & < 0.0001\\
    Grounding & 1.32 & 13.344 & < 0.0001\\
    Gratitude & 1.14 & 11.463 & < 0.0001\\
    Agreement & 1.33 & 13.317 & < 0.0001\\
    \bottomrule
    \end{tabular}
    \caption{Mean difference between each strategy and the paraphrase baseline, Wald test statistics and p-values for the receptiveness experiment contrasts. The test is conducted for each strategy compared to paraphrased reply.}
    \label{tab:statistics_receptiveness_contrasts}
\end{table}

\begin{table*}[!t]
    \centering
    \begin{tabular}{ccccc}
    \toprule
    \textbf{Strategy} & \textbf{Pair} & \textbf{Mean diff} & \textbf{Test stat.} & \textbf{p-value} \\
    \midrule
    Avg. Strategies & high - med &  0.146 & 1.567 & 0.266\\
    & high - low & $\textbf{0.385}^{**}$ & 4.107 & 0.0003\\
    & med - low & $\textbf{0.239}^*$ & 2.540 & 0.0342\\
    \midrule
    Paraphrase (B) & high - med &  0.094 & 0.497 & 0.873\\
    & high - low & 0.273 & 1.458 & 0.312\\
    & med - low & 0.181 & 0.934 & 0.619\\
    \midrule
    Receptive (B) & high - med &  -0.0734 & -0.457 & 0.891\\
    & high - low & 0.0530 & 0.331 & 0.941\\
    & med - low & 0.126 & 0.788 & 0.711 \\
    \bottomrule
    \end{tabular}
    \caption{Mean difference between each strategy and the paraphrase baseline, Wald test statistics and p-values for the toxicity-receptiveness contrasts. The test is conducted for the receptiveness score at each toxicity level compared to the others. ``B'' indicates a baseline. Bolded values are significant. $^*$ indicates significance at p < 0.05, $^{**}$ indicates significance at p < 0.001 }
    \label{tab:statistics_toxicity_contrasts}
\end{table*}

\section{Annotation Interfaces}
\label{appendix:interfaces}
This section includes the annotation interfaces used for all the annotation tasks in this paper.
\begin{figure*}[!t]
    \centering
    \includegraphics[width=\textwidth]{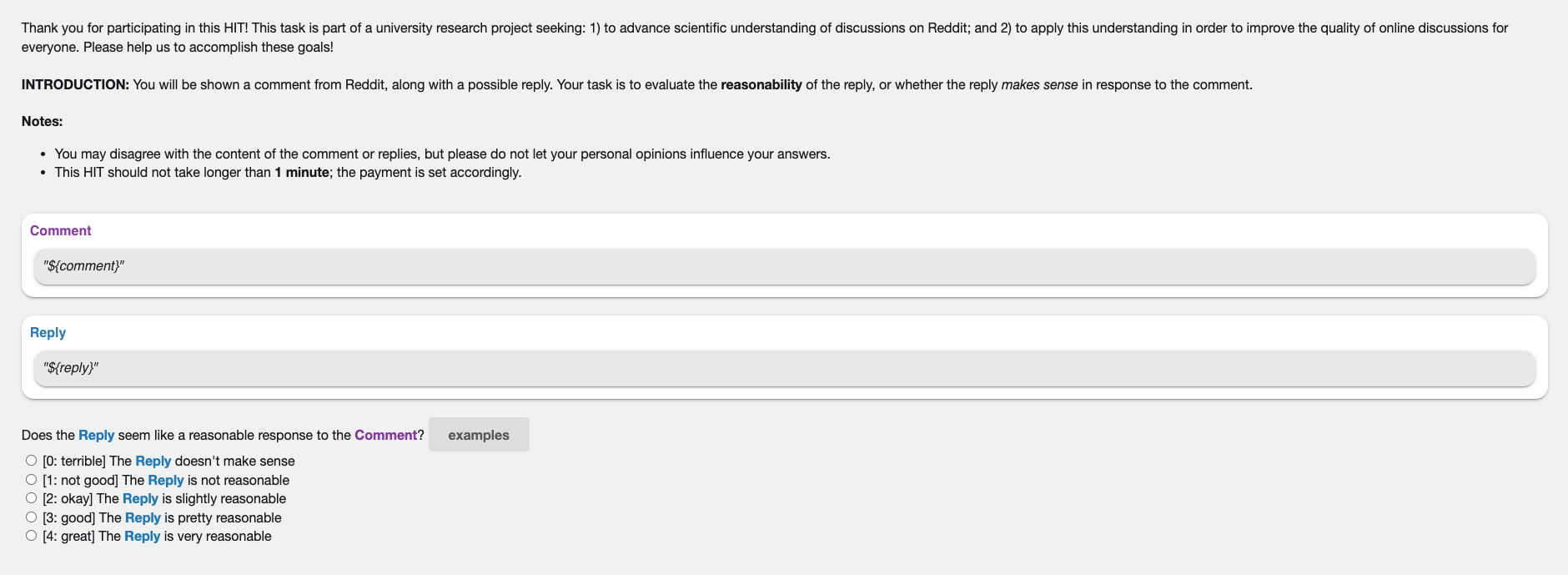}
    \caption{Reasonability validation task interface}
    \label{fig:plausibility-interface}
\end{figure*}

\begin{figure*}[!t]
    \centering
    \includegraphics[width=\textwidth]{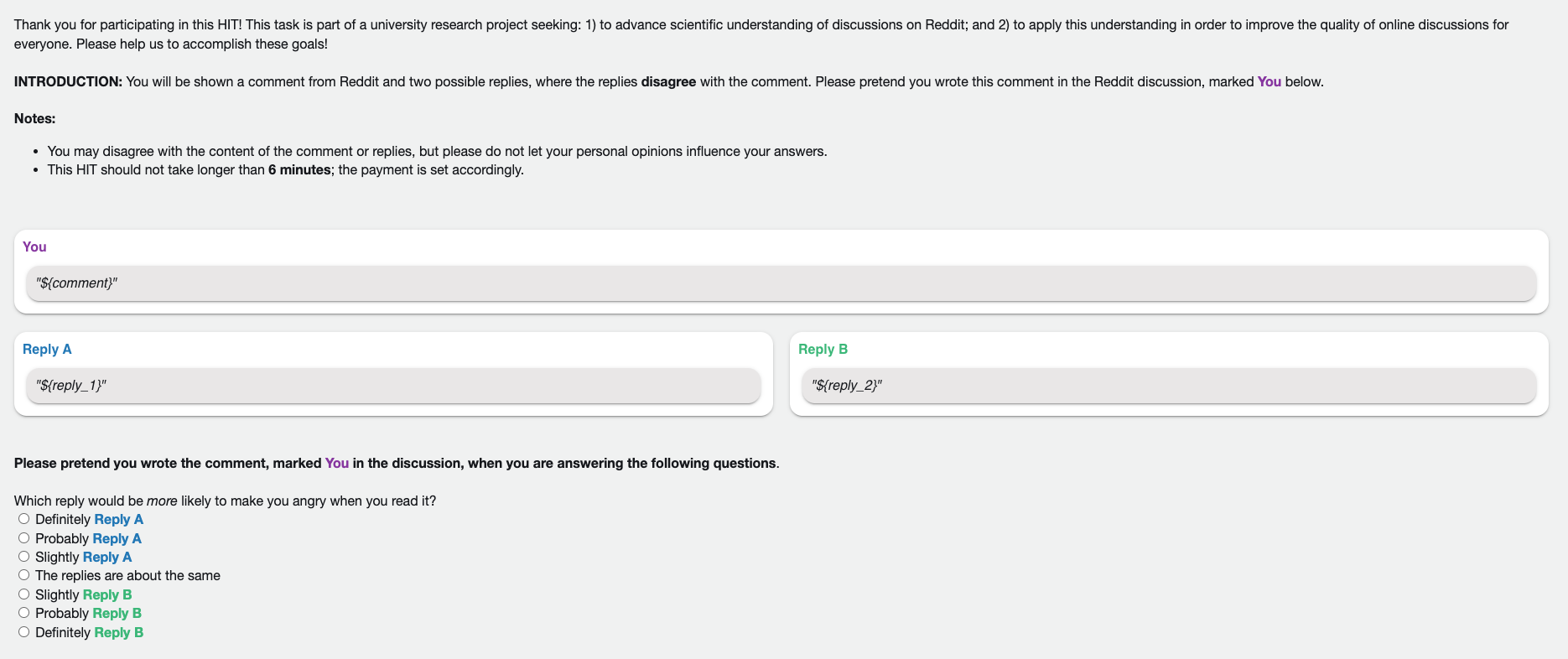}
    \caption{Receptiveness task interface (a)}
    \label{fig:receptiveness-interface-1}
\end{figure*}

\begin{figure*}[!t]
    \centering
    \includegraphics[width=\textwidth]{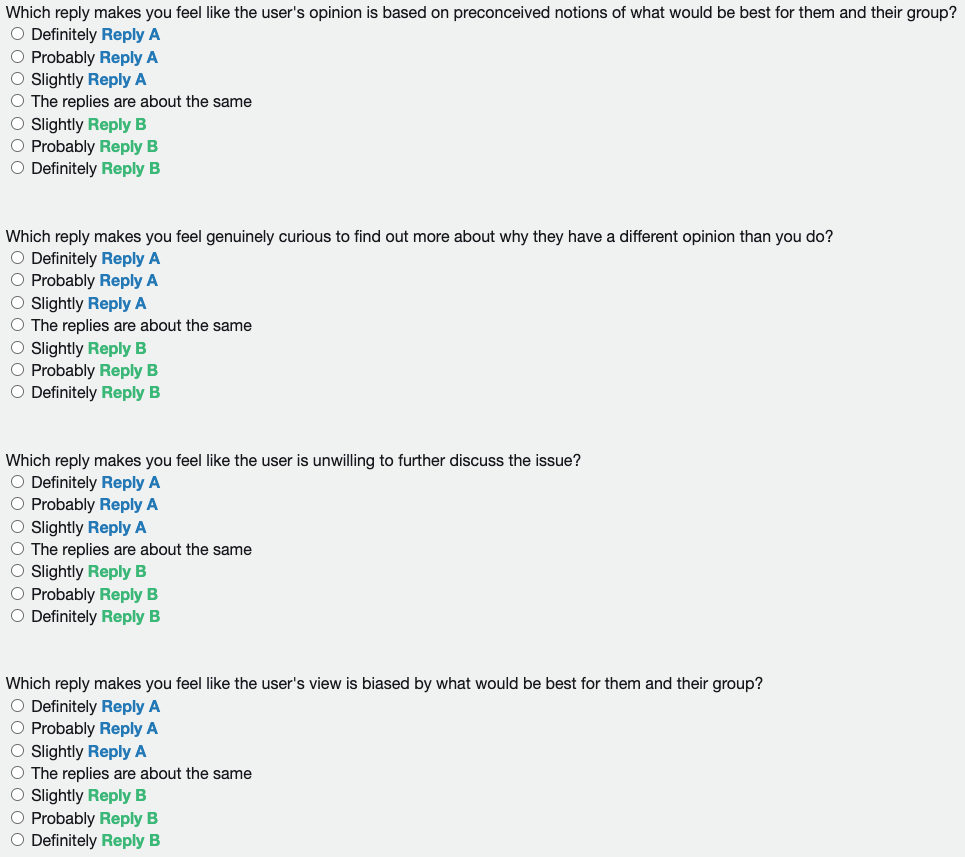}
    \caption{Receptiveness task interface (b)}
    \label{fig:receptiveness-interface-2}
\end{figure*}

\begin{figure*}[!t]
    \centering
    \includegraphics[width=\textwidth]{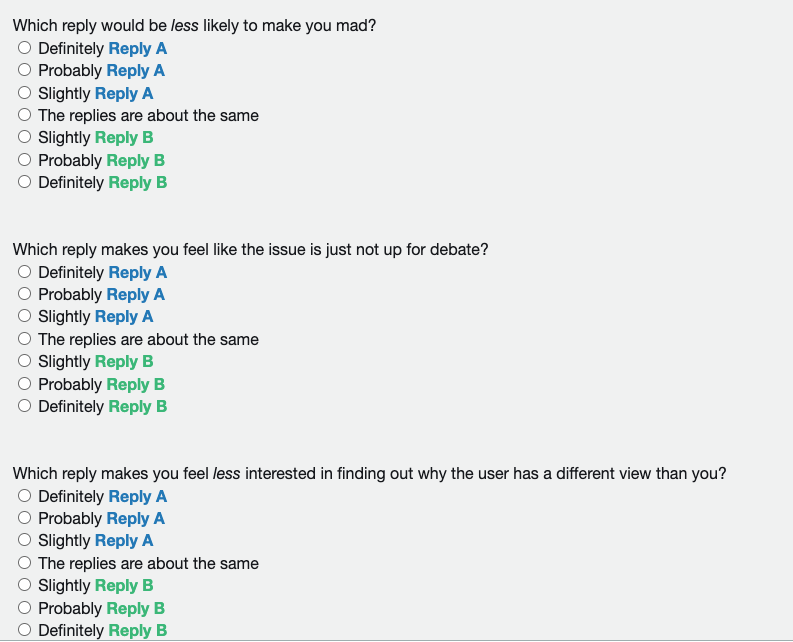}
    \caption{Receptiveness task interface (c)}
    \label{fig:receptiveness-interface-3}
\end{figure*}

\end{document}